\documentclass[10pt,twocolumn,letterpaper]{article}

\usepackage{iccv}
\usepackage{times}
\usepackage{epsfig}
\usepackage{graphicx}
\usepackage{amsmath}
\usepackage{amssymb}

\usepackage{cuted}
\usepackage{comment}
\usepackage{overpic}
\usepackage{capt-of}
\usepackage[pagebackref=true,breaklinks=true,letterpaper=true,colorlinks,bookmarks=false]{hyperref}

\iccvfinalcopy 
\newcommand{\tablestyle}[2]{\setlength{\tabcolsep}{#1}\renewcommand{\arraystretch}{#2}\centering\footnotesize}

\newcounter{mysfig}
\counterwithin{mysfig}{figure}
\renewcommand\themysfig{\thefigure(\alph{mysfig})}
\newcommand\themysfigcite{(\alph{mysfig})}
\makeatletter
\newcommand\Scaption[1]{%
\addtocounter{figure}{1}
\refstepcounter{mysfig}%
\vskip.5\abovecaptionskip
  \sbox\@tempboxa{\footnotesize\themysfigcite~#1}%
  \ifdim \wd\@tempboxa >\hsize
    \small\themysfig~#1\par
  \else
    \global \@minipagefalse
    \hb@xt@\hsize{\hfil\box\@tempboxa\hfil}%
  \fi
  \vskip\belowcaptionskip
  \addtocounter{figure}{-1}
}

\makeatother

\ificcvfinal\fi

\begin{document}

\title{Learning a Room with the Occ-SDF Hybrid: \\Signed Distance Function Mingled with Occupancy Aids Scene Representation}

\author{
Xiaoyang Lyu$^{1}$,
~Peng Dai$^{1}$,
~Zizhang Li$^{2}$,
~Dongyu Yan$^{3}$,
~Yi Lin$^{4}$,
~Yifan Peng$^{1}$,
~Xiaojuan Qi$^{1}$,
\vspace{0.8em}\\
$^{1}$The University of Hong Kong , $^{2}$  Zhejiang University,  $^{3}$  Harbin Institute of Technology,  $^{4}$ DJI
}

\maketitle

\begin{strip}

     \centering
     \includegraphics[width=0.95\linewidth]{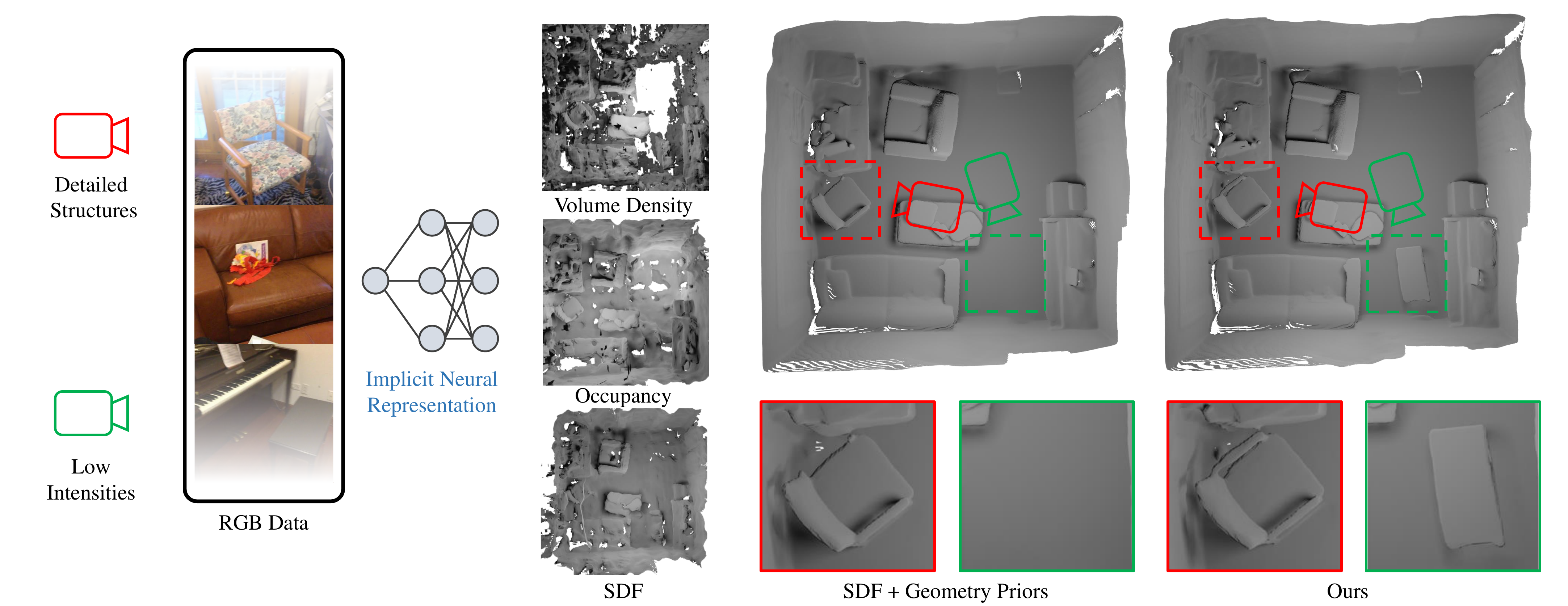}
     \captionof{figure}{\textbf{Overview and reconstruction results of the Occ-SDF hybrid neural scene representation.} Aided with the feature rendering scheme (Sec.~\ref{sec: feature}) and hybrid representation (Sec.~\ref{sec: hybrid}), our method yields results with more detailed structures in room-level scenes compared to the state-of-the-art, especially for those low intensities and detailed structures.
     }
     \label{fig:teaser}
\end{strip}
\begin{abstract}
Implicit neural rendering, which uses signed distance function (SDF) representation with geometric priors (such as depth or surface normal), has led to impressive progress in the surface reconstruction of large-scale scenes. However, applying this method to reconstruct a room-level scene from images may miss structures in low-intensity areas or small and thin objects. We conducted experiments on three datasets to identify limitations of the original color rendering loss and priors-embedded SDF scene representation.

We found that the color rendering loss results in optimization bias against low-intensity areas, causing gradient vanishing and leaving these areas unoptimized. To address this issue, we propose a feature-based color rendering loss that utilizes non-zero feature values to bring back optimization signals. Additionally, the SDF representation can be influenced by objects along a ray path, disrupting the monotonic change of SDF values when a single object is present. To counteract this, we explore using the occupancy representation, which encodes each point separately and is unaffected by objects along a querying ray. Our experimental results demonstrate that the joint forces of the feature-based rendering loss and Occ-SDF hybrid representation scheme can provide high-quality reconstruction results, especially in challenging room-level scenarios. The code would be released.
\end{abstract}

\section{Introduction}
\label{sec:intro}

Reconstructing a 3D scene from a series of multi-view images is a crucial problem in the realm of computer vision. This process has widespread applications in various fields such as animation, gaming, and virtual/augmented reality (VR/AR). 
The recent trend is to represent a 3D scene as an implicit function parameterized by a neural network \cite{Mescheder_2019_CVPR, park2019deepsdf, sitzmann2020metasdf, peng2020convolutional}, whose optimization is supervised by explicit 3D data like point cloud or real SDF value.
Recent advancements in neural radiance field (NeRF) \cite{mildenhall2021nerf} further enable learning an implicit 3D representation from purely sparse posed images \cite{yu2021pixelnerf,mueller2022instant}.

However, when it comes to producing high-quality novel-view synthesis, these methods frequently utilize volume density~\cite{mildenhall2021nerf} to represent the 3D geometry. 
Unfortunately, this approach does not adequately constrain the 3D geometry in the presence of ambiguities~\cite{oechsle2021unisurf}, ultimately leading to poor surface reconstructions (as depicted in Fig.~\ref{fig:teaser}: Volume Density).

Accordingly, research efforts have been made to exploit geometry-friendly representations, including signed distance function (SDF)~\cite{wang2021neus, yariv2021volume, park2019deepsdf} or occupancy~\cite{oechsle2021unisurf}, whose zero-level set can be  extracted to become the concerned 3D surface. Albeit improving quality, they consider the reconstruction only of a single object, thereby, the performance degrades dramatically when applied to scene-level surface reconstruction, {\ie}, representing a room ({Fig.~\ref{fig:teaser}: SDF}).  
An attribute is that reconstructing texture-less areas often suffers from ambiguous visual cues with only RGB loss as the regularization. 
To address this problem, recent research has attempted to incorporate semantic \cite{guo2022neural} or geometric priors (depth/normal~\cite{yu2022monosdf, wang2022neuris} constraints) to further regularize scene-level reconstruction. 
With SDF-based representation and geometric priors~\cite{yu2022monosdf}, the reconstruction quality has been greatly improved ({Fig.~\ref{fig:teaser}}: SDF + Geometry Priors), especially concerning large flat areas and objects. However, it still cannot faithfully reconstruct the 3D scene with missing structures in low-intensity dark areas and small/thin objects ({Fig.~\ref{fig:teaser}}: SDF + Geometry Priors). 

The above observation motivates us to dive into bridging the remaining missing blocks of existing neural surface representation methods. 
Notably, we focus on the SDF-based representation as it achieves state-of-the-art performance and has been widely adopted. 
Our analysis suggests that both the RGB color rendering formulation and SDF representation have clear limitations preventing existing solutions from fully unleashing the potential of implicit neural surface representation for large-scale room-level scenes.  

First, the color itself can show a significant impact on the optimization of geometric representation relying on the original RGB-based rendering formula \cite{mildenhall2021nerf}, namely color bias. 
In particular, dark pixels with small intensity values will make the partial derivation of the loss with respect to the corresponding SDF value become zero, corrupting the optimization and resulting in missing structures in dark areas (see for example {in Fig.~\ref{fig:teaser}: Low Intensities}). 
Accordingly, herein instead of directly calculating the weighted color, we first compute weighted features and then use a learnable multi-layer perceptron (MLP) to decode the final rendering color. 
In such a way, we would still be able to effectively optimize the corresponding geometry representation as long as the feature vector contains non-zero values.

Second, the vanilla SDF-based neural rendering only considers a single ray directly passing through the object surface from the empty space and ignores objects along the ray \cite{wang2021neus, yariv2021volume}. 
This configuration violates scene-level geometry where the existence of multiple objects clearly affects the distributions of SDF (Fig.~\ref{fig:toy_3D}). 
Meanwhile, the optimization of thin structures and small objects, which naturally has small sampling probability, will be greatly degraded by this violation even with correct geometry prior {and the structure will be erased to minimize the global geometry loss (Fig.~\ref{fig:teaser}: Detailed Structures)}.


In addition, although occupancy-based representations are likely to generate unwanted structure and cannot warrant a smooth surface reconstruction ({Fig.~\ref{fig:teaser}}), they are often sufficiently robust to objects along the ray and free from object interference in scene-level data. Therefore, during optimization, we propose to describe the room-level scene using occupancy in conjunction with signed distance functions (SDFs) to compensate for each other's defects. 

The technical contributions are as follows:

\begin{itemize}
    \item We explore an improved feature rendering scheme to overcome the problem of vanishing gradients in neural implicit reconstruction brought by the vanilla color space rendering formula.
    \item  We carefully investigate insights and limitations in existing SDF and occupancy representations, and accordingly propose a hybrid representation mingling SDF with occupancy, dubbed \textbf{Occ-SDF Hybrid}, to resolve surfaces with thin structures and small objects.
    \item We conduct a large body of qualitative and quantitative experiments against state-of-the-art, indicating that our Occ-SDF hybrid formula can yield a higher-fidelity room-level scene representation, particularly with successfully resolving small and dark objects.
\end{itemize}

\section{Related Work}
\label{related works}

\noindent\textbf{Multi-view Stereo} Conventional algorithms~\cite{agrawal2001probabilistic, schoenberger2016mvs, dellaert2000structure} always split the reconstruction into two steps. First, the feature-matching method \cite{rublee2011orb, ng2003sift, schonberger2016pixelwise} is applied to estimate the depth of each frame. Then, the resulting depth maps~\cite{merrell2007real} are used to reconstruct the final scene. 
Notably, the reconstruction may suffer from poor performance in texture-less areas. Learning-based approaches are mainly divided into two categories.
Typically, neural networks are embedded into the traditional reconstruction pipeline to replace specific modules, like feature matching \cite{sarlin2020superglue, zagoruyko2015learning, ummenhofer2017demon}, depth estimation \cite{yao2018mvsnet}, or depth fusion \cite{donne2019learning}.
These methods often suffer from depth inconsistency problems due to the separately estimated depth maps.
Alternatively, neural networks are designed to directly regress input images to truncated signed distance functions (TSDFs) 
 \cite{murez2020atlas, sun2021neuralrecon}, but the reconstruction results often lack enough fine details.

\noindent\textbf{Neural Scene Representation} Recently, coordinate-based neural representations can faithfully model a 3D scene with only posed images. 
Approaches with an implicit differentiable renderer \cite{yu2021pixelnerf} only use volume density as scene representation which can not extract 3D scenes directly. 
To address this issue, occupancy-based representation \cite{oechsle2021unisurf} and SDF-based representation \cite{wang2021neus, yariv2021volume} are proposed to facilitate 3D reconstruction. 
Notably, these methods already achieve great performance for object-level scenes but exhibit poorly for room-level scenes, especially in texture-less areas.

\noindent\textbf{Priors for Indoor Scene Reconstruction} Existing methods have attempted to introduce priors to resolve higher-fidelity surfaces in texture-less areas. 
Manhattan-SDF \cite{guo2022neural} follows semantic-NeRF\cite{Zhi:etal:ICCV2021} to estimate the volume density and semantic label at the same time, and then uses Manhattan-World assumption to regularize the geometry in floor and wall regions. 
NeuRIS \cite{wang2022neuris} and MonoSDF \cite{yu2022monosdf} directly exploit the depth and normal predicted from an off-the-shelf neural network to regularize the geometry of each point, but in many cases, thin structures would disappear.
NeuRIS \cite{wang2022neuris} proposes a dynamic scheme to eliminate the wrong supervision signal from inaccurate estimated results, however, based on our investigation that fine structures are still lost even with the correct geometry for supervision. 
In all, we seek to explore a feature rendering scheme and a hybrid representation to overcome the above problems.

\section{Overview and Preliminary}
\label{sec: overview}
Our goal is to examine the \emph{limitations} of existing implicit neural surface representations and explore practical \emph{solutions} for accurately reconstructing large-scale, room-level 3D geometry with fine details from a set of calibrated images. 
First,  we find that the well-adopted color-based rendering formula in \cite{yu2022monosdf, wang2022neuris}  will induce optimization bias against low-intensity areas, leaving these areas under-optimized and resulting in missing reconstructions ({Sec.~\ref{subsec: rendering_problem}}). 
Accordingly, we propose a simple yet effective feature-based rendering formula to address the problem ({Sec.~\ref{subsec: feature_rendering}}).
Second, our analysis shows that the SDF-based neural surface representation violates scene-level geometry supervised signal and thus prevents the model from obtaining accurate reconstructions, making the model tends to sacrifice small and thin structures ({Sec.~\ref{subsec: geo_problem}}). 
Motivated by this, {we propose a hybrid representation mingling occupancy and SDF for accurate reconstruction} ({Sec.~\ref{subsec: hybrid_representation}}). 

We here describe the mathematical preliminary on the state-of-the-art surface representation, namely {SDF-based Neural Scene Representation}~\cite{yariv2021volume}, for 3D reconstruction. 

For implicit neural reconstruction, we can represent the scene as a signed distance function (SDF) field, which is a continuous function $f$ that calculates the distance between each point and its closest surface
\setlength\abovedisplayskip{3.0pt}
\setlength\belowdisplayskip{3.1pt}
\begin{equation}
    \begin{aligned}
        1_\Omega(\boldsymbol{p})&=\left\{
        \begin{array}{ll}
            1 & \text { if } \boldsymbol{p} \in \Omega \\
            0 & \text { if } \boldsymbol{p} \notin \Omega
        \end{array}, \right. \\
         f(\boldsymbol{p})&=(-1)^{1_\Omega(\boldsymbol{p})} \min _{\boldsymbol{y} \in \mathcal{M}}\|\boldsymbol{p}-\boldsymbol{y}\| \;,
    \end{aligned}
\end{equation}
where $1_\Omega(\boldsymbol{p})$ is an indicator function to represent whether the space at position $p$ is occupied, $\mathcal{M} = \partial\Omega$ is the boundary surface of occupied space and $||\cdot||$ is the standard Euclidean 2-norm. 
    Following the VolSDF~\cite{yariv2021volume}, we use an MLP to represent the function $f$ and convert the SDF value to Laplace density with the following function
\begin{equation}
    \sigma_i(\boldsymbol{p}_i)=\alpha \Psi_\beta\left(-f(\boldsymbol{p}_i)\right) \;,
\end{equation}
where $\alpha, \beta > 0$ are learnable parameters, and $\Psi_\beta$ is the cumulative distribution function (CDF) with zero mean and the $\beta$ scale is defined as
\begin{equation}
    \Psi_\beta(s)= \begin{cases}
    \frac{1}{2} \exp \left(\frac{s}{\beta}\right) &, \text { if } s \leq 0 \\ 1-\frac{1}{2} \exp \left(-\frac{s}{\beta}\right) &, \text { if } s>0
    \end{cases} .
\end{equation}
\begin{figure}[t]
     \centering
     \begin{overpic}[width=0.94\linewidth]{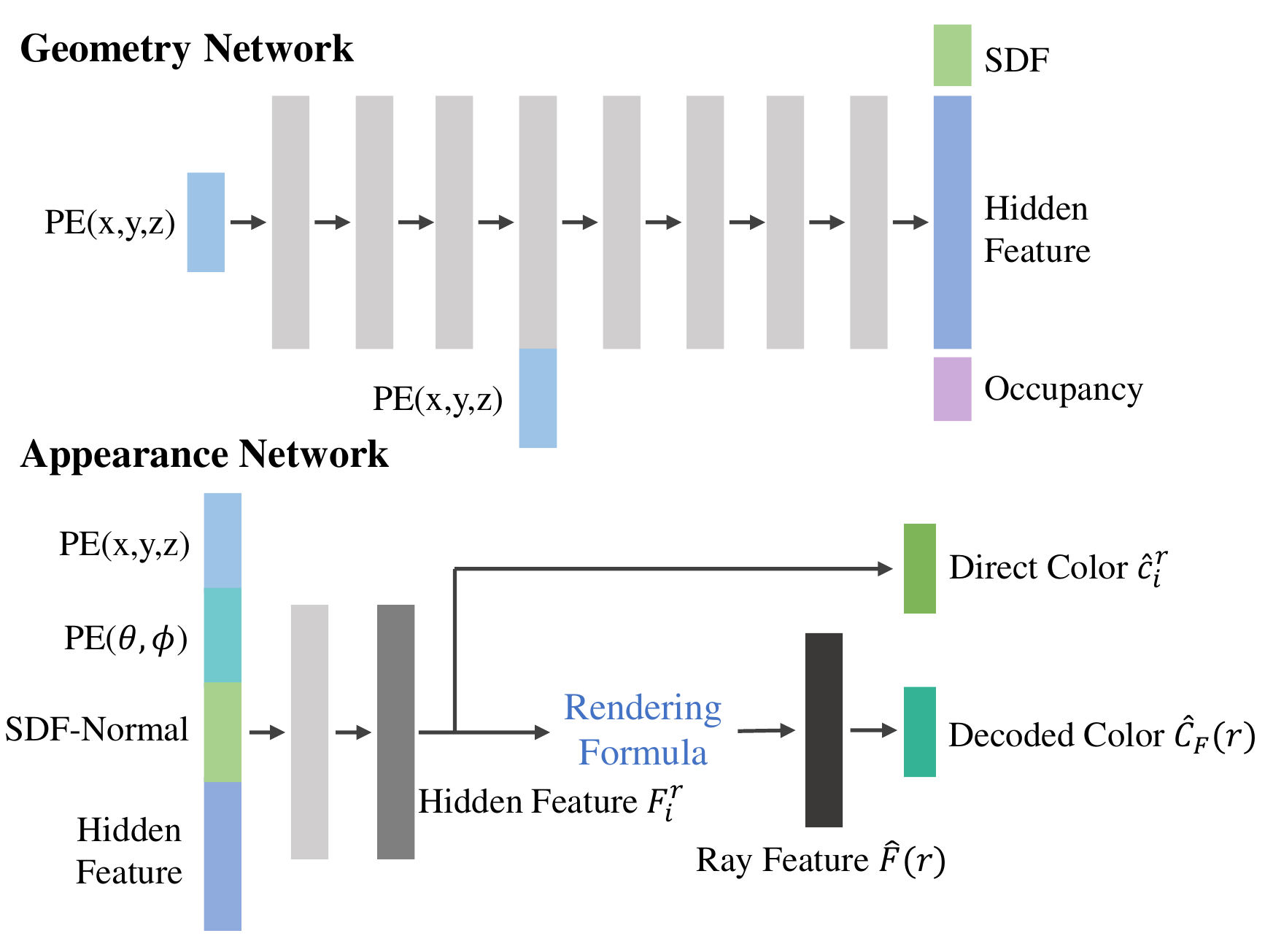}
     \footnotesize
     \put(47, 21.5){\textcolor[rgb]{0.27, 0.45, 0.77}{Eq.~\eqref{eq: feature rendering}}}
     \end{overpic}
     \caption{\textbf{Network architecture.} 
     The geometry network takes 3D position $(x, y, z)$ after positional encoding(PE) as input and output both SDF and occupancy value. The appearance network takes view direction $(\theta, \phi)$ as input and outputs two types of color, the direct color is used in Eq. \eqref{eq: color_rendering} to directly obtain the pixel color and the decoded color is calculated via the rendering formula (Eq.~\eqref{eq: feature rendering}). 
     }
     \label{fig:method}
     \vspace{-16pt}
 \end{figure}

\noindent\textbf{Color Rendering Formula}
According to the rendering formula~\cite{mildenhall2021nerf}, the color for the current ray $r$ is rendered by
\begin{equation}
\label{eq: color_rendering}
    \hat{C}(r) = \sum_{i=1}^{M}T^r_i\alpha_i\hat{c}^r_i \;,
\end{equation}
where $T_i^r$ and $\alpha_i$ represent the transmittance and alpha value (a.k.a opacity), respectively, of sampled point. They can be computed by
\begin{equation}
    T^r_i = \prod_{j=1}^{i-1}(1-\alpha_i), ~
    \alpha_i = 1 - \exp(-\sigma^r_i\delta^r_i) \;,
\end{equation}
where $\delta_r^i$ is the distance between adjacent sample points.
Finally, given the rendered color $\hat{C}(r)$,  the SDF field will be optimized from sparsely sampled images by minimizing the color-based rendering loss as
\begin{equation}
\label{eq: c_loss}
    \mathcal{L}_\text{rgb} = \sum_{r\in R}||\hat{C}(r) - C(r)||_1 \;,
\end{equation}
where $C(r)$ is the ground-truth color associated with the sampled ray $r$.

\section{Feature Rendering Formula}
\label{sec: feature}


\subsection{Problem of Color-based Rendering}
\label{subsec: rendering_problem}

Given the SDF-based scene representation and the color-based rendering loss $\mathcal{L}_\text{rgb}$ in {Sec.~\ref{sec: overview}}, we analyze the derivative of $\mathcal{L}_\text{rgb}$ to the opacity $\alpha_i$ of a point $p_i$. 
Note that for a single point $p_i$, as $\alpha^r_i$ are the same regardless of the rays, we thus omit its dependency on ray $r$ and use $\alpha_i$ for simplicity.
For a point $\boldsymbol{p}_i$, the derivative of the color loss function to its opacity $\alpha_i$ is
\begin{equation}
\label{eq: derivate}
    \frac{\partial \mathcal{L}_{rgb}}{\partial \alpha_i} \!\!=\!\! \pm \!\!\left(\prod_{j=1}^{i-1}(1-\alpha_j)c_i - \!\!\!\!\!\sum_{k=i+1}^N\!\!\!\!c_k\alpha_k\!\!\!\!\prod_{j=1, j\ne i}^{k-1}\!\!\!(1-\alpha_j)\right)\!\!,
\end{equation}
\noindent which indicates that when we optimize the SDF value of $\boldsymbol{p}_i$, the gradient is determined by the color of the current point $\boldsymbol{p}_i$ and points behind it ($c_k$ and $k \in\{i+1,..., N\}$), and opacity of all points on the entire ray except for the current point ($\alpha_j$ and $j\in\{1, 2,...N\}~\&~j\neq i$).
Notably, when processing a dark region, saying that the $c_i$ approaches zero, the first term of Eq.~\eqref{eq: derivate} will be close to zero. 
{Similarly}, if points behind {$\boldsymbol{p}_i$} have low opacity ($\alpha_k$ and $k \in\{i+1,..., N\}$), the gradient with respect to the SDF value will be small, causing the vanishing problem in dark regions. 
More generally, the gradient of SDF values can be affected by the color itself, resulting in a biased optimization process that tends to favor high color intensities.

\begin{figure}[t]
  \centering
    \begin{minipage}[t]{0.28\linewidth}
    \begin{minipage}[c]{\linewidth}
        \includegraphics[width=\linewidth]{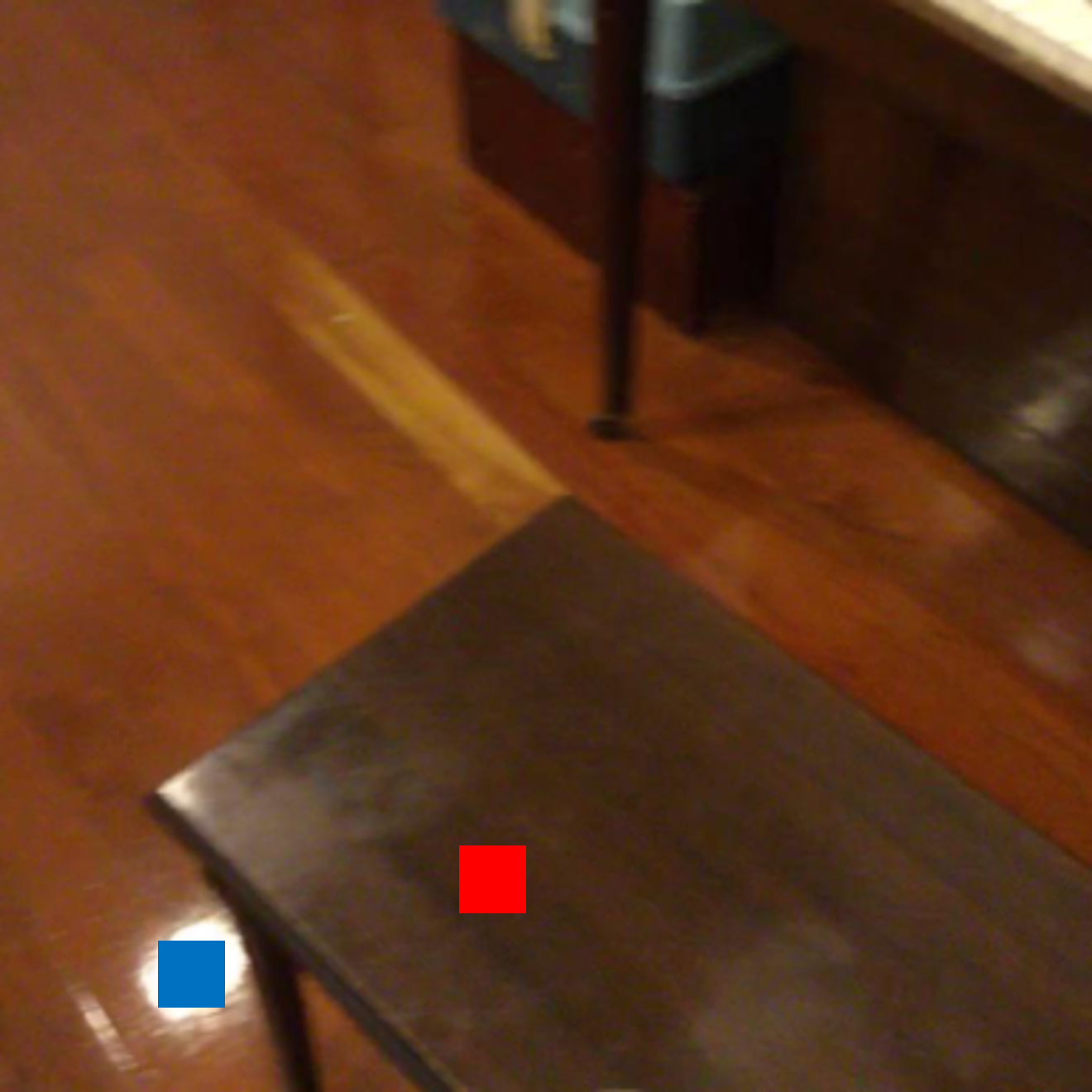}
    \end{minipage} 
    \end{minipage}
    \begin{minipage}[t]{0.65\linewidth}
    \begin{minipage}[c]{\linewidth}
        \includegraphics[width=\linewidth]{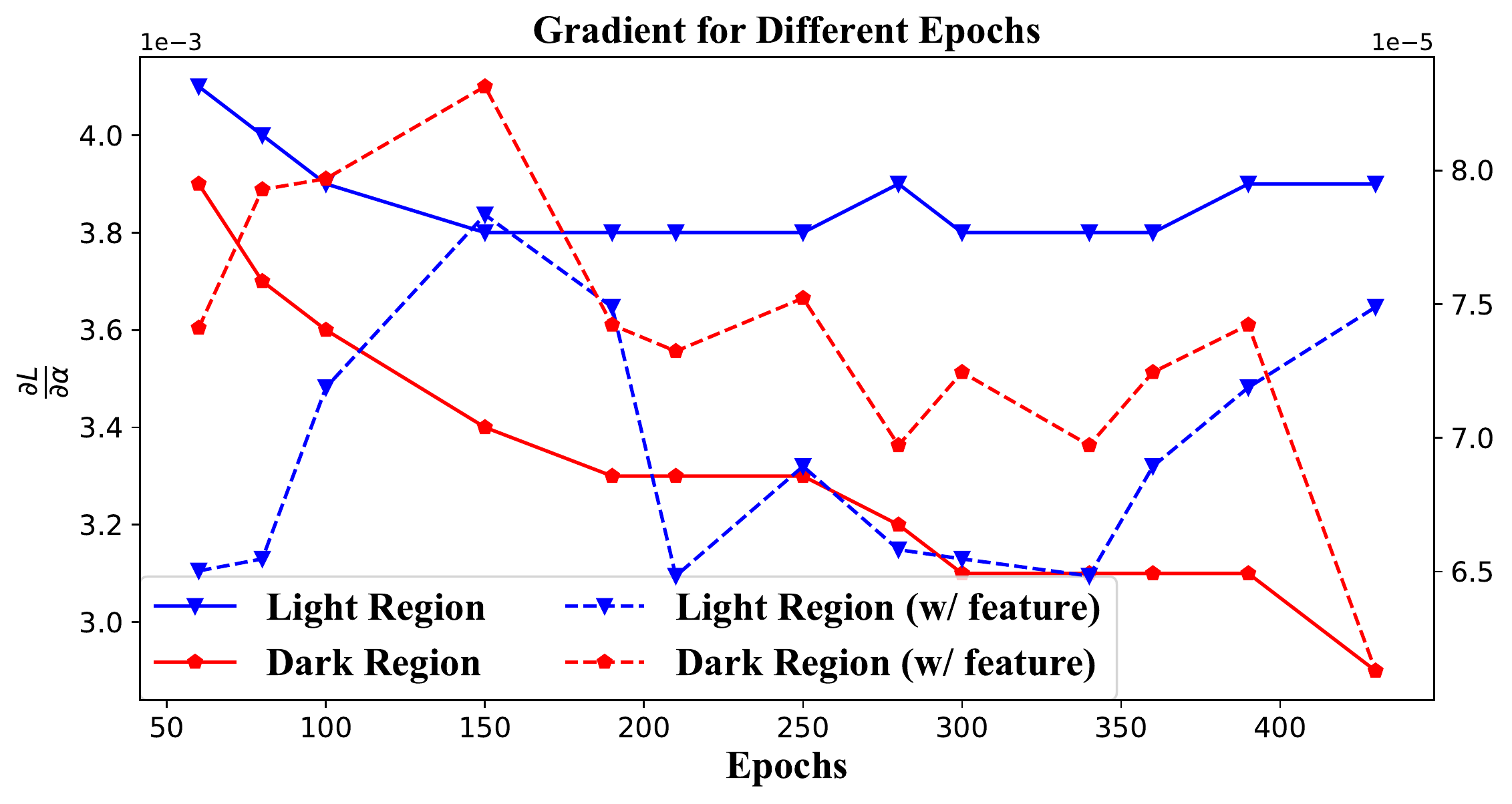}
    \end{minipage}
    \end{minipage}
    
    \caption{\textbf{Verification of the color influence.} {Left: Rays sampled in the dark region (red block) and the light region (blue block). Right: The trend of gradient norm corresponding to two regions during optimization.}} 
    \label{fig: feature rendering scheme}
    \vspace{-14pt}
\end{figure}

The above analysis is also supported by our experiments below. As shown in {Fig.~\ref{fig: feature rendering scheme}}, we sample rays in the dark region (\textcolor{red}{red} square) and the light region (\textcolor{blue}{blue} square) separately and accordingly record the trend of gradient norms in these two regions during the optimization.
In the beginning, the gradient norms from these two regions are similar; and the gradient norm in the dark region (\textcolor{red}{red} solid line) decreases as the number of training epochs increases, while the gradient norm in the light region remains stable (\textcolor{blue}{blue} solid line), indicating the dark areas contribute much less to the optimization process. As the optimization process proceeds, if these points are predicted as dark colors ($c^r_i \rightarrow 0$), it would lead to the gradient reduction effects as analyzed in Eq.~\eqref{eq: derivate}. 
Note that the gradient of points in these areas will not equal zero due to the influence of other loss functions, like depth consistency loss.

\subsection{Feature-based Rendering}
\label{subsec: feature_rendering}

{To resolve the aforementioned problem, we propose feature-based color rendering loss. 
{As shown in Fig.~\ref{fig:method},} the \emph{Appearance} network outputs two predictions for each point $i$ along a ray $r$: one is the color vector $\hat{c}_i^r$, and the other is the hidden feature $F^r_i$. 
For direct color $\hat{c}_i^r$, we utilize Eq.~(\ref{eq: color_rendering}) to obtain the target pixel color $\hat{C}_{c}(r)$. 
And the hidden feature $F_r^i$ is used to render the ray feature $\hat{F}(r)$ by
\begin{equation}
\label{eq: feature rendering}
    \hat{F}(r) = \sum_{i=1}^MT^r_i\alpha_iF_i^r \;.
\end{equation}
The ray feature $\hat{F}_r$ is further decoded by a decoder $\mathcal{D}$ to yield the decoded target pixel color, 
\begin{equation}
\label{eq: decode feature}
    \hat{C}_F(r) = \mathcal{D}(\hat{F}(r)) \;,
\end{equation}
where the decoder $\mathcal{D}$ is a single-layer perceptron with 256 nodes. 
Finally, the decoded color $\hat{C}_F(r)$ from the rendered feature is used to construct the feature-based color rendering loss. 
As such, the optimization of these dark regions would not be affected by the color itself. As long as there are non-zero values in the rendered feature, there will be non-zero gradients with respect to the volume density of the concerned point. {As shown in Fig.~\ref{fig: feature rendering scheme} (dashed lines), the gradient norm is not influenced by the intensity of colors. } }
\section{Hybrid Representation Scheme}
\label{sec: hybrid}

{\subsection{Incorporating Geometry Prior Matters}}

It is clear that for room-level scene reconstruction, geometry priors are essential.
Existing methods~\cite{wang2022neuris, yu2022monosdf} render depth $\hat{D}(r)$ and normal $\hat{N}(r)$ of the surface intersecting the current ray as
\begin{equation}
\label{eq: n_d_represent}
    \hat{D}(r) = \sum_{i=1}^{M}T^r_i\alpha^r_it^r_i\quad \text{and} \quad 
    \hat{N}(r) = \sum_{i=1}^{M}T^r_i\alpha^r_i\hat{n}^r_i \;,
\end{equation}
where $\hat{T}_r^i$ and $\hat{\alpha}_r^i$ have the same meaning as Eq.~\eqref{eq: color_rendering}, $t_i^r$ is the distance the ray passing and $n_i^r$ is the normal of point $p_i$.
Next, these methods use depth and normal maps estimated from pre-trained models, such as Omnidata \cite{kar20223d}, to directly supervise the rendered depth $\hat{D}(r)$ and normal $\hat{N}(r)$ using Eq.~\eqref{eq: d_loss} and Eq.~\eqref{eq: n_loss}, respectively.
Overall, the depth loss function is defined as
{\setlength\abovedisplayskip{3pt}
\setlength\belowdisplayskip{3pt}
\begin{equation}
\label{eq: d_loss}
    \mathcal{L}_\text{depth} = \sum_{r\in\mathcal{R}}||(w\hat{D}(r) + q) - \bar{D}(r)||^2 \;, 
\end{equation}
}
where $w$ and $q$ are the scale and shift computed by the least-squares method~\cite{eigen2014depth} to solve scale-ambiguity problem for monocular depth prediction methods. 
And the normal loss function is
{\setlength\abovedisplayskip{3pt}
\setlength\belowdisplayskip{3pt}
\begin{equation}
\label{eq: n_loss}
    \mathcal{L}_\text{normal}\!\! =\!\! \sum_{r\in R}\!\!||\hat{N}(r) - \bar{N}(r)||_1\!\! +\!\! ||1 - \hat{N}(r)^T\bar{N}(r)||_1 \;,
\end{equation}
}
where $\bar{N}(r)$ is the predicted monocular normal transformed to the same coordinate system with angular.

As shown in Fig.~\ref{fig:teaser}, we note that these geometry priors benefit the reconstruction of better surfaces in texture-less and sparse-viewed areas.
However, thin structures and small objects, such as the yellow flower in Fig.~\ref{fig: disappear}, {cannot be faithfully reconstructed with geometry priors. Recently, NeuRIS~\cite{wang2022neuris} put forward a hypothesis that  this phenomenon arises from the inaccurate geometry supervisory signal ({\ie} depth and surface normal). However, according to our experiment on Replica synthetic dataset,  this problem still exists even though we use the perfect ground-truth depth, normal, and RGB to provide supervisory signals (see Fig.~\ref{fig: disappear}). 

\subsection{Problem of SDF Formula with Geometry Prior}
\label{subsec: geo_problem}

To dive into the SDF representation and explore its limitations for surface reconstruction with geometric priors, 
we create a simplified scenario and simulate object occlusions as shown in Fig.~\ref{fig:toy_3D}. 
The ground-truth SDF intersecting with a horizontal plane is shown in Fig.~\ref{fig:toy_3D}, and the SDF distribution along a ray $r$ intersecting with the blue cube at point $p_{s}$ is shown in Fig.~\ref{fig:toy_sdf_ray}.
There are many local minima and maxima due to the existence of multiple objects, which differs from the single-object scenario following a monotonic function (blue line in Fig.~\ref{fig:toy_sdf_ray}). 
To examine depth priors for surface reconstruction, we employ the approach proposed in MonoSDF~\cite{yu2022monosdf} to calculate the depth of $p_{s}$ following Eq.~\eqref{eq: n_d_represent} subject to the ground-truth SDF. 

However, even with the ground-truth SDF, the estimated depth value (1.59, the red vertical plot in Fig.~\ref{fig:toy_weights_ray}) still deviates from the true depth value (2.94, the green vertical plot in Fig.~\ref{fig:toy_weights_ray}) when multiple objects exist. 
According to Eq.~\eqref{eq: n_d_represent}, the estimated normal value would suffer from the same problem.
This implies that existing methods incorporating {geometry priors \cite{yu2022monosdf, wang2022neuris}} to guide the learning of the SDF representation may not necessarily encourage the model to learn the true SDF for scene-level surface reconstruction. 
In turn, because small objects or thin structures usually have low sampling probability during training, the minimization of $\mathcal{L}_\text{depth}$ will encourage the model to predict SDF ignoring small objects along the ray $r$ such that the estimated SDF will produce depth values closer to the depth supervision (see Fig.~\ref{fig:toy_weights_ray}) and minimize the overall loss function, attempting to mimic the single object scenario (right part in Fig.~\ref{fig:toy_experiment}). 
In sum, the supervision from geometric priors tends to sacrifice the reconstruction of small objects to preserve large surface reconstruction, which aligns also with our observation presented in Fig.~\ref{fig: disappear}. 

\begin{figure}[t]
  \centering
   \includegraphics[width=0.98\linewidth]{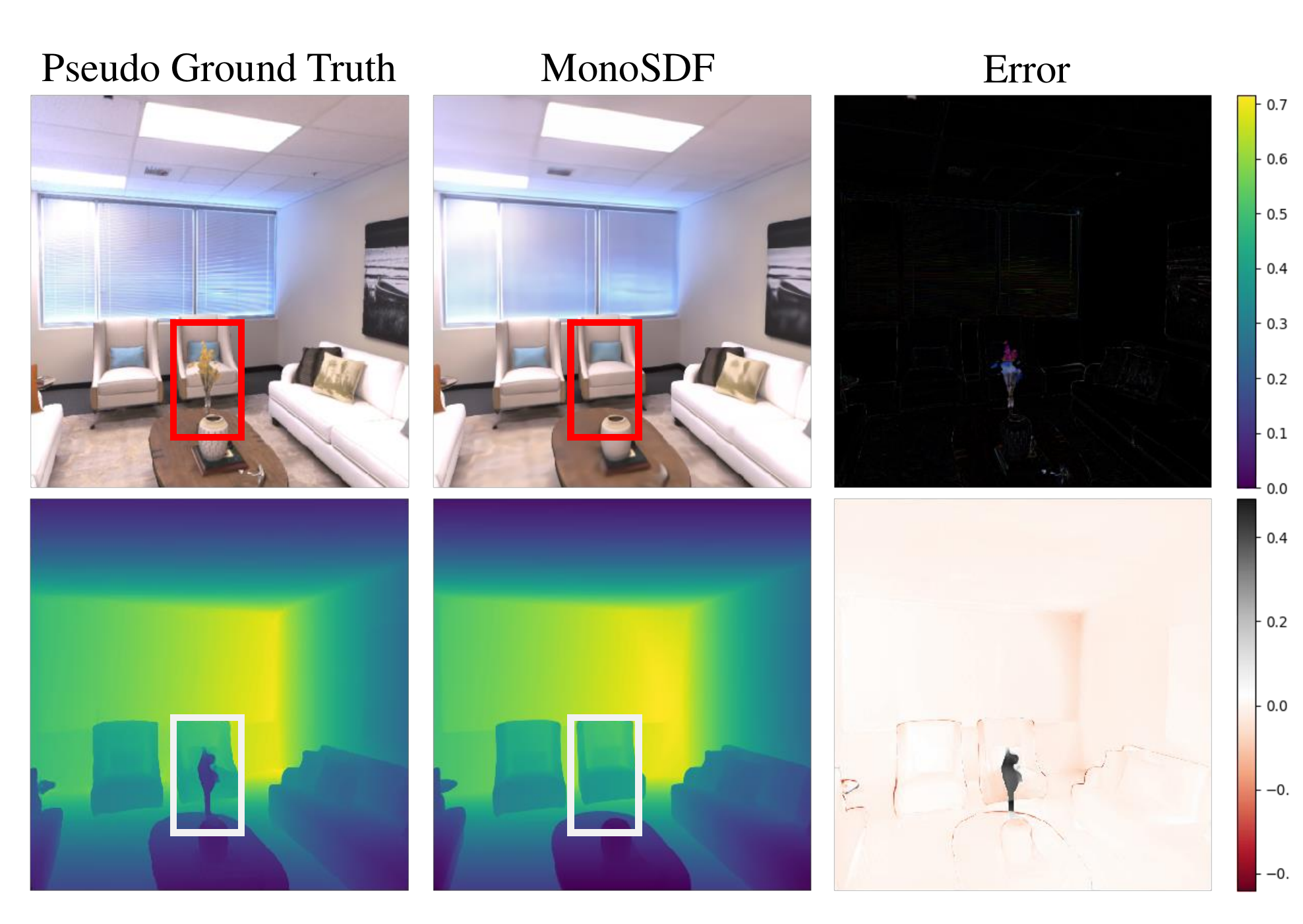}
   \vspace{-6pt}
   \caption{\textbf{Illustration of failure cases of state-of-the-arts.} Even though applying the perfect pseudo ground-truth geometry to supervise the model, existing room-level reconstruction methods like~\cite{yu2022monosdf} can still fail to resolve accurate 3D structures. 
   }
   \label{fig: disappear}
   \vspace{-12pt}
\end{figure}

\subsection{Hybrid Occupancy-SDF Representation 
}
\label{subsec: hybrid_representation}

The problem above is essentially caused by the SDF representation, which describes the geometry of a scene as a whole and thus suffers from the interference of other objects, especially small objects, and thin structures are prone to be removed to preserve large structures.  

Unlike the SDF representation, occupancy represents each point separately and thus is free from the interference of objects in this challenging scenario. 
However, occupancy representation only focuses on the intersecting point ignoring the constraint of neighborhood points. Thus, the reconstruction results represented by occupancy will have many floaters and useless structures, as shown in Fig.~\ref{fig:teaser}, which can be eliminated in SDF by Eikonal loss \cite{gropp2020implicit}.
This inspires us to investigate a hybrid of occupancy and SDF as a representation for neural surface reconstruction as shown in Fig.~\ref{fig:method}. 
The geometry network $\theta$ outputs both SDF  and occupancy. 

Specifically, the occupancy represents surfaces as the decision boundary of a binary occupancy classifier, parameterized by a neural network $\theta$
\begin{equation}
    o_{\theta}(\textbf{p}): \mathbb{R}^3\rightarrow [0, 1] \;,
\end{equation}
where $\textbf{p}$ is a 3D point.
The occupancy representation assumes that objects are solid, thus we can rewrite the neural rendering formula~\cite{mildenhall2021nerf} to
\begin{equation}
    \hat{C}(r)=\sum_{i=1}^N o(\textbf{p}_i)\prod_{j<i}(1 - o(\textbf{x}_j))c(\textbf{p}_i, d) \;,
\end{equation}
which replaces the opacity $\alpha$ to a discrete occupancy indicator variable $o\in{0,1}$, where $o = 0$ indicates the free space while $o = 1$ the occupied space. Thus, this representation will not be affected by the objects along the ray, and the rendering depth will be consistent with ground-truth depth in ground-truth occupancy space. {Note that the occupancy-based representation is introduced to facilitate optimization. During inference, the SDF is used for reconstruction. }



{To understand why and how the hybrid representation helps optimize the SDF field for accurate reconstruction, we conduct the following empirical analysis, using the scenario shown in Fig.~\ref{fig:toy_3D}. 
Here, the orange ray hits a surface point $Q$ of the large blue cube and the blue ray hits the surface point $P$ on the small cylinder. 
During optimization, the depth/normal loss for point $Q$ along the orange ray will encourage the model to predict a large SDF value (absolute) of point P (Fig.~\ref{fig:sdf point P}) which violates the reconstruction of the small cylinder where a small SDF value is desired. 
In contrast, point $Q$ has no effects on point $P$ with  occupancy representation.
The hybrid representation joins the forces of SDF and occupancy representations, aiming to use occupancy representation to help overcome the issues of SDF representation in optimization. 
Although the depth/normal loss from the SDF presentation for point $Q$ still has a negative impact on the optimization of point $P$. 
The additional occupancy representation will force the network to predict a large occupancy value for a point $P$ and thus will indirectly regularize the network to predict a small SDF value (see Fig.~\ref{fig:sdf point P}: the blue up arrow in ``Hybrid'').
We admit that this hybrid representation can only alleviate this problem, and our study is more empirical. 
Fundamental issues arise from insufficient neural scene representation, which requires further research efforts. 
We explore further why this combination would bring notable benefits to the supplement with an example.}


\begin{figure*}
\vspace{-5pt}
  \centering
    \begin{minipage}[h]{0.6\paperwidth}
    \begin{minipage}[b]{\linewidth}
        \includegraphics[width=0.6\paperwidth]{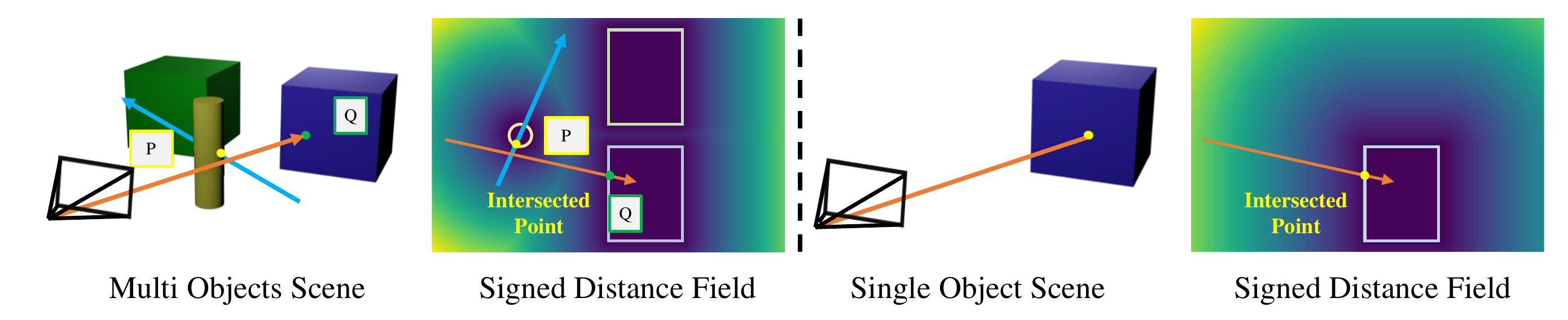}
        \vspace{-18pt}
        \Scaption{Schematic of the toy 3D structure for the multi-object scene (left) and single object scene (object). }
        \label{fig:toy_3D}
    \end{minipage}
    \end{minipage}
    \begin{minipage}[h]{0.16\paperwidth}
    \begin{minipage}[b]{\linewidth}
        \includegraphics[width=\linewidth]{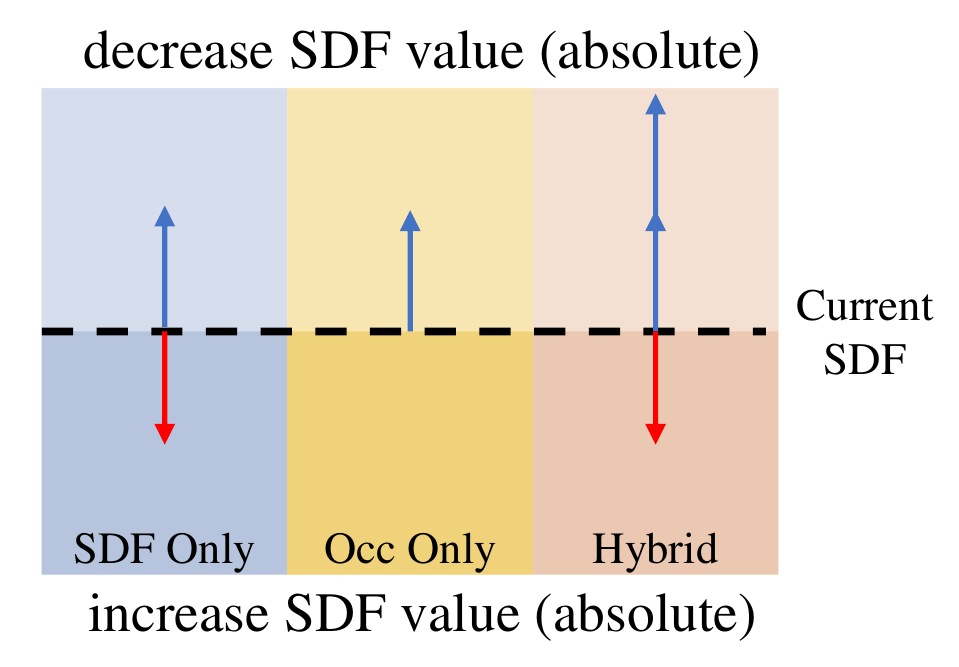}
        \vspace{-18pt}
        \Scaption{SDF Value at Point P}
        \label{fig:sdf point P}
    \end{minipage}
    \end{minipage}
    \begin{minipage}[h]{0.22\paperwidth}
    \begin{minipage}[b]{\linewidth}
        \includegraphics[width=\linewidth]{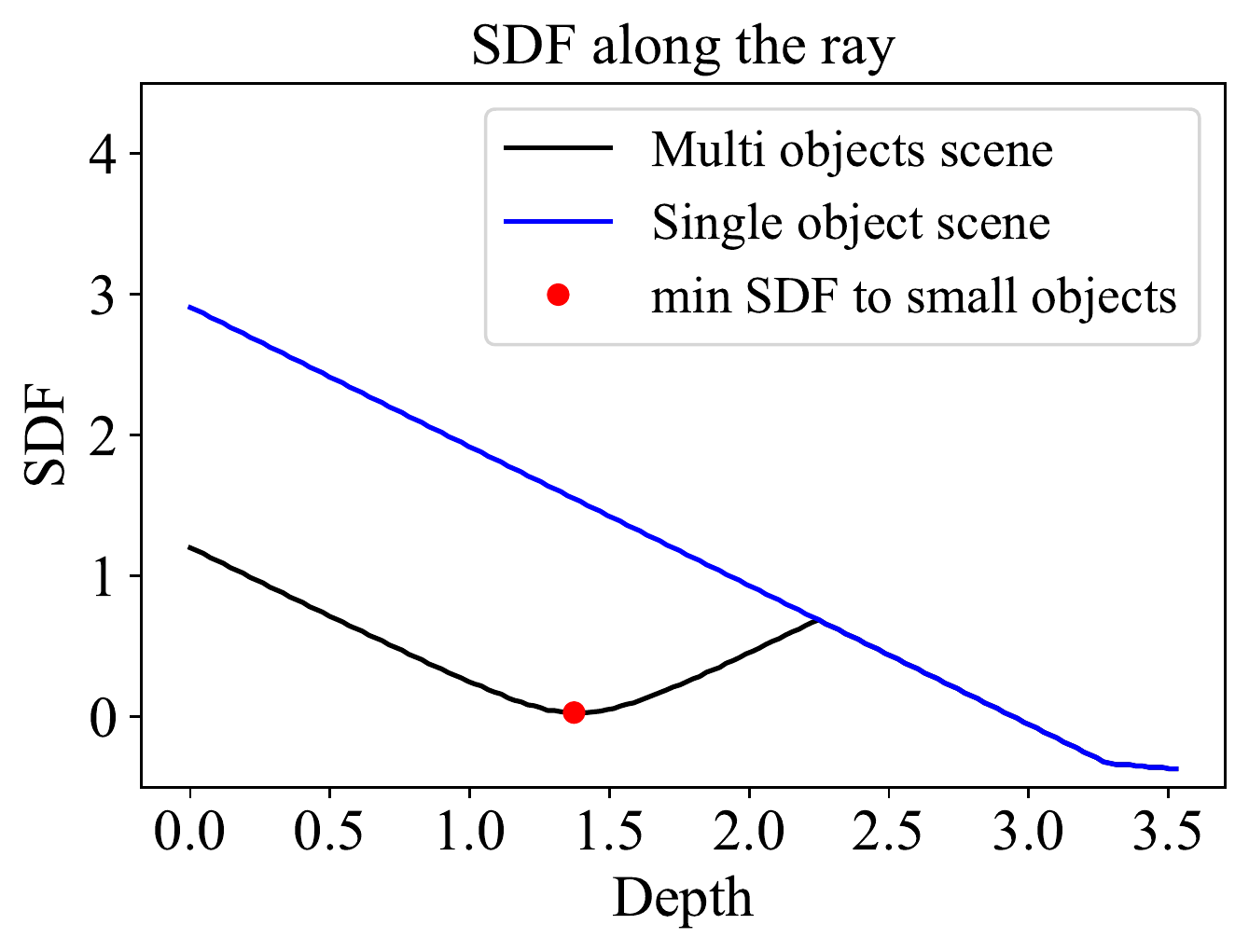}
        \vspace{-18pt}
        \Scaption{SDF  $f(\boldsymbol{p}_i)$}
        \label{fig:toy_sdf_ray}
    \end{minipage}
    \end{minipage}
    \begin{minipage}[h]{0.22\paperwidth}
    \begin{minipage}[b]{\linewidth}
        \includegraphics[width=\linewidth]{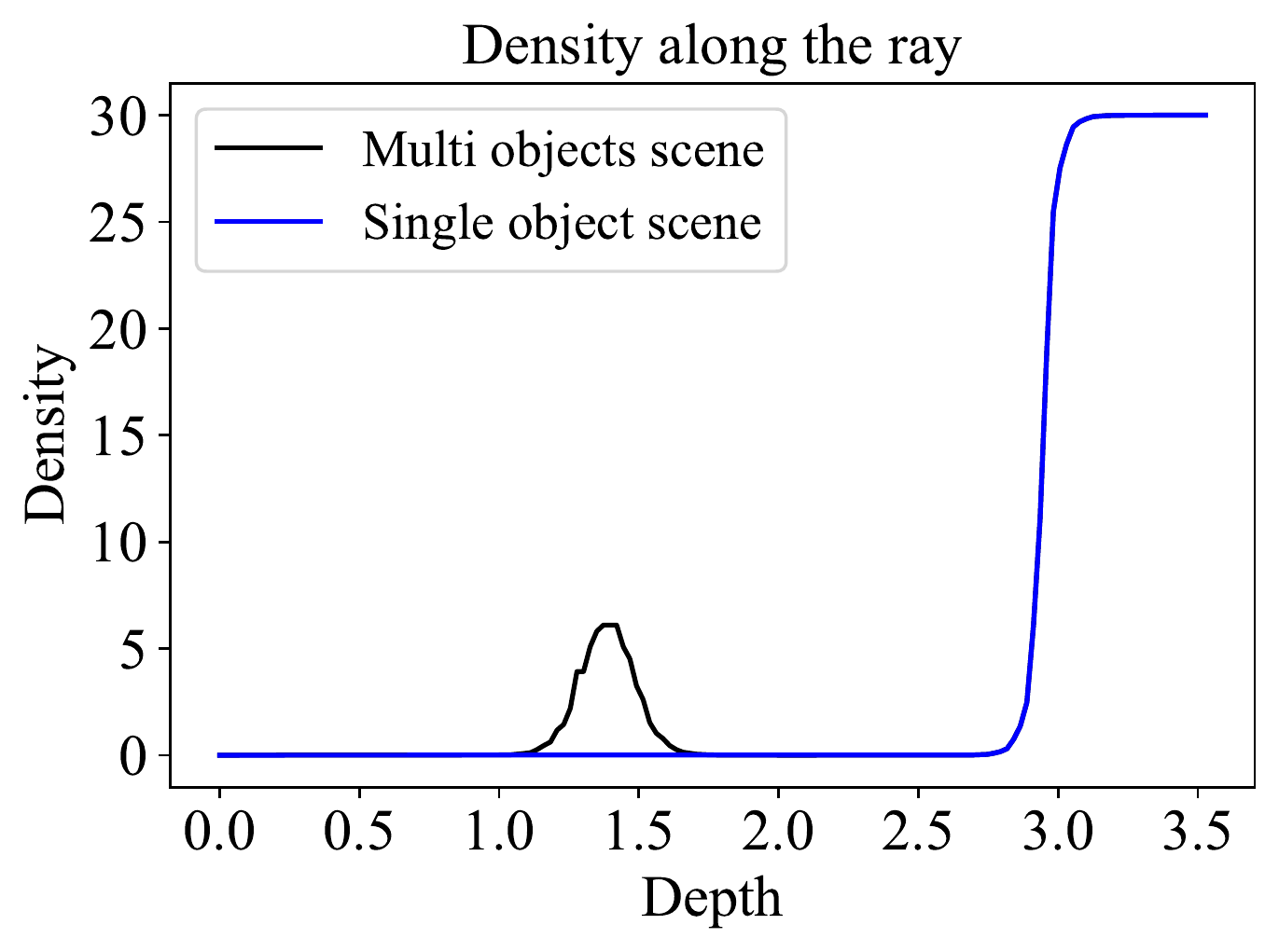}
        \vspace{-18pt}
        \Scaption{Laplace density  $\delta_i(\boldsymbol{p}_i)$}
        \label{fig:toy_density_ray}
    \end{minipage}
    \end{minipage}
    \begin{minipage}[h]{0.34\paperwidth}
    \begin{minipage}[b]{\linewidth}
        \includegraphics[width=\linewidth]{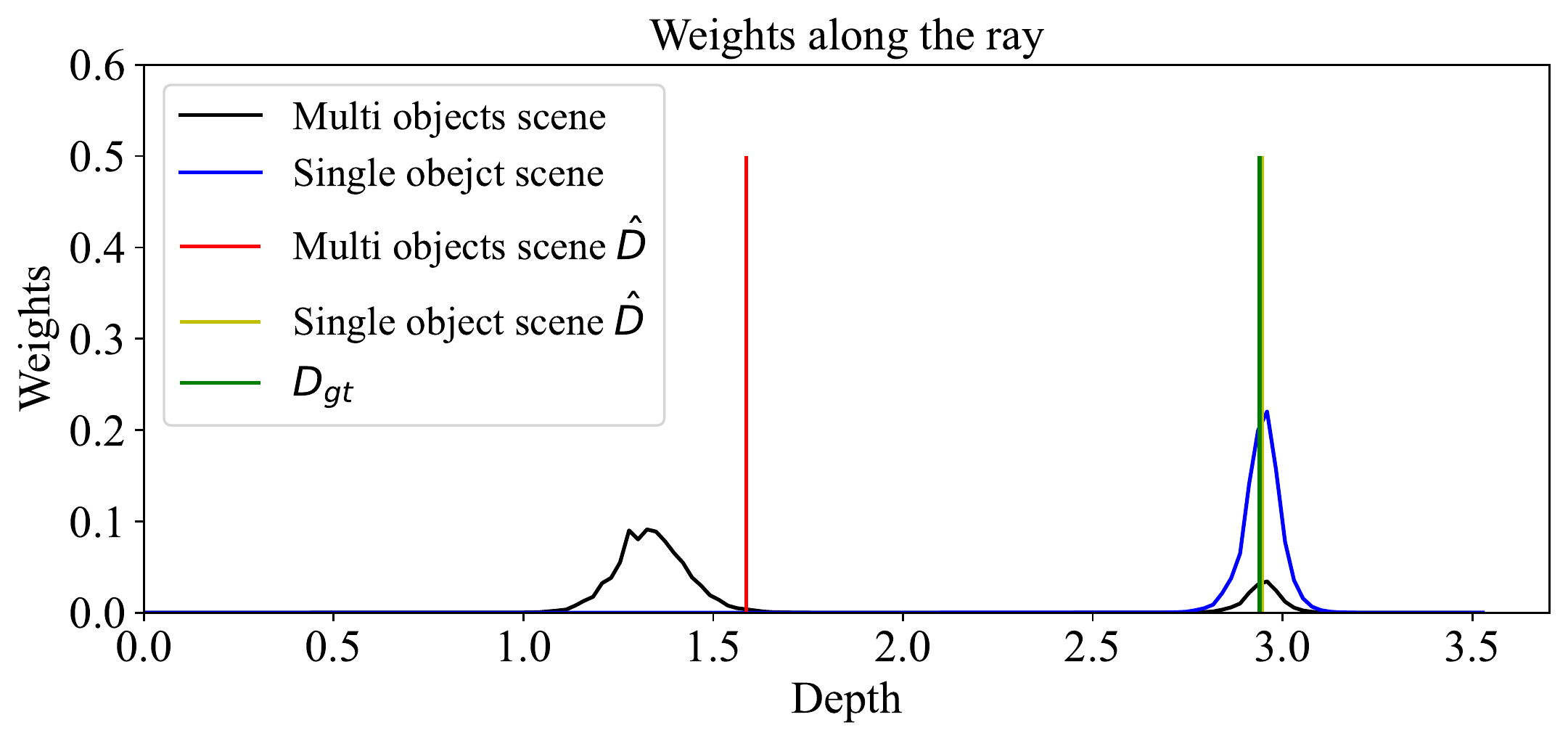}
        \vspace{-18pt}
        \Scaption{Weights $T_r^i\alpha_r^i$}
        \label{fig:toy_weights_ray}
    \end{minipage}
    \end{minipage}
    
    \vspace{3pt}
    \caption{\textbf{Toy experiments for the room-level scene.} The left part of \ref{fig:toy_3D}, where we stand in front of the yellow cylinder to observe the entire scene, is a widespread scenario for room-level scale scenes. 
    Unlike the single object scenario, where the distribution of SDF value is a monotonic decreasing function from the observed position to the object, the room-level scenario has complex distributions with multi peaks/valleys along the single ray~(\ref{fig:toy_sdf_ray}). 
    Following the Laplace density function~\cite{yariv2021volume}, the density distributions of different situations are shown in \ref{fig:toy_density_ray}, where room-level scenes have a secondary peak near small objects but the single object scene only has one peak. 
    It is because of the existence of this peak, the weights in the room-level scene~(\ref{fig:toy_weights_ray}) exhibit a multi-model distribution, while for the single object case a uni-modal distribution.
    As such, we note that the rendering depth $\hat{D}$ deviates from the ground truth in the room-level scene but is close to the ground truth depth object-level scene. \ref{fig:sdf point P} means the effect of supervised signal in three different representations.}
    \label{fig:toy_experiment}
    \vspace{-9pt}
\end{figure*}

\section{Experiments}
\label{sec:exp}
\noindent\textbf{Optimization.}
In the training stage, we minimize the loss
{\setlength\abovedisplayskip{6pt}
\setlength\belowdisplayskip{6pt}
\begin{equation}
\begin{aligned}
\label{eq: overall_loss}
    \mathcal{L} = &\mathcal{L}^{sdf_F}_\text{rgb} + \mathcal{L}^{sdf}_\text{rgb} + \lambda_1 \mathcal{L}_\text{eik} \\ 
    &+ \lambda_2 \mathcal{L}_\text{depth}^\text{occ} + \lambda_3 \mathcal{L}_\text{depth}^\text{sdf} + \lambda_4 (\mathcal{L}_\text{normal}^{occ} + \mathcal{L}_\text{normal}^{sdf}),
\end{aligned}
\end{equation}
}
\noindent where $\mathcal{L}_{rgb}$ means color-based rendering loss following Eq.~\eqref{eq: c_loss} but only to SDF representation. The rendering color computed by the feature rendering formula (Eq.\eqref{eq: feature rendering} and Eq.~\eqref{eq: decode feature}) is denoted to $\mathcal{L}_{rgb}^{sdf_F}$.
Notably, $\mathcal{L}_{eik}$ means the eikonal loss~\cite{gropp2020implicit}, 
$\mathcal{L}_{depth}$ means the depth rendering loss following Eq.~\eqref{eq: d_loss}, and $\mathcal{L}_{normal}$ means the normal rendering loss following Eq.~\eqref{eq: n_loss}.
We apply normal and depth loss for both representations, where the superscript $^{occ}$ indicates the loss computed by occupancy-based representation, while the $^{sdf}$ is by SDF-based representation.
The network is optimized by the Adam optimizer with a learning rate of $5e^{-4}$. 
We set the weights $\lambda_1, \lambda_2, \lambda_3, \lambda_4$ to $0.05, 1, 0.1, 0.05$, respectively. 
The network architecture and sampling strategy are detailed in the supplement.

\vspace{0.05in} \noindent\textbf{Datasets.} We use three datasets to assess the performance of our algorithm. \textbf{ScanNet}~\cite{dai2017scannet} is a real-world dataset that provides 1,513 scenes captured with Kinect V1 RGB-D camera. The BundleFusion~\cite{dai2017bundlefusion} is applied to provide high-quality camera poses and surface reconstructions. For each scene, we uniformly sample roughly 500 frames to train our network.
{\textbf{Tanks and Temples}~\cite{Knapitsch2017} is a real-world, large-scale scene dataset. We use four indoor scenes from their advanced split and run on the official server. 
} \textbf{Replica}~\cite{replica19arxiv} is a synthetic dataset that provides 18 scenes, with each providing dense geometry, HDR textures, and semantic annotations. We select 8 scenes and use the Habitat simulator~\cite{szot2021habitat} to render RGB images following MonoSDF~\cite{yu2022monosdf} splits. 
Notably, we conduct {ablation studies} on this dataset. 

\vspace{0.05in}
\noindent\textbf{Compared Methods.} 
(1) \textbf{UNISURF}~\cite{oechsle2021unisurf} is an occupancy-based method that unifies surface rendering and volume rendering for neural scene reconstruction. We implement the \textbf{UNISURF*} with normal and depth priors for a fair comparison. (2) \textbf{MonoSDF}~\cite{yu2022monosdf} is  an SDF-based method that adds depth and normal constraints on  VolSDF~\cite{yariv2021volume}.
(3) \textbf{Manhattan-SDF}~\cite{guo2022neural} is an SDF-based method that adds a semantic branch and uses the Manhattan constraint to regularize the geometry in floor and wall regions.
(4) \textbf{COLMAP}~\cite{schoenberger2016mvs} is a classical multi-view stereo method with  Poisson surface reconstruction. 
(5) \textbf{NeuRIS}~\cite{wang2022neuris} is an SDF-based method that introduces pseudo normal prior to the NeUS~\cite{wang2021neus} architecture. Meanwhile, it leverages multi-view consistency to eliminate the wrong supervision signal from inaccurate estimated results.}

\vspace{0,05in}
\noindent\textbf{Metrics.} 
All meshes are evaluated by 5 standard metrics defined in~\cite{murez2020atlas}: \emph{Accuracy}, \emph{Completeness}, \emph{Precision}, \emph{Recall}, and \emph{F-score}. 
Their definition will be discussed in the supplementary.
For the Replica dataset, we also report the normal consistency following~\cite{Mescheder_2019_CVPR, guo2022neural, Peng2021SAP}.
For the Tanks and Temples dataset, we use the official server to evaluate our results and report the F-score for selected scenes.

\subsection{Main Results}

We compare our method with state-of-the-art methods on three benchmark datasets. 

\begin{table}[h]
  \centering
  \footnotesize
   \setlength\tabcolsep{2pt}
  \begin{tabular}{l|cccccc}
    \hline
    Method & Acc $\downarrow$  & Comp $\downarrow$ & C-$\mathcal{L}_1$ $\downarrow$ & Prec $\uparrow$ & Recall $\uparrow$ & F-score $\uparrow$ \\
    \hline
    COLMAP \cite{schoenberger2016mvs}        & 0.047    & 0.235 &  0.141 & 71.1 & 44.1 & 53.7\\
    UNISURF \cite{oechsle2021unisurf}       & 0.554    & 0.164 &  0.359 & 21.2 & 36.2 & 26.7\\
    VolSDF \cite{yariv2021volume}        & 0.414    & 0.120 &  0.267 & 32.1 & 39.4 & 34.6\\
    NeUS \cite{wang2021neus}          & 0.179    & 0.208 &  0.194 & 31.3 & 27.5 & 29.1 \\
    Manhattan-SDF \cite{guo2022neural} & 0.072    & 0.068 &  0.070 & 62.1 & 56.8 & 60.2\\
    NeuRIS \cite{wang2022neuris} & 0.050    & 0.049 &  0.050 & 71.7 & 66.9 & 69.2\\
    MonoSDF \cite{yu2022monosdf}    & \textbf{0.035}    & 0.048 &  0.042 & 79.9 & 68.1 & 73.3\\
    \textbf{Ours} & 0.039    & \textbf{0.041} &  \textbf{0.040} & \textbf{80.0} & \textbf{76.0} & \textbf{77.9}\\
    \hline
  \end{tabular}
  \vspace{2pt}
  \caption{Quantitative assessments of the proposed model against previous works on the ScanNet dataset. }
  \vspace{-12pt}
  \label{tab: scannet}
\end{table}

\begin{figure*}[t!]
    \vspace{-12pt}
    \centering
    \includegraphics[width=0.97\linewidth]{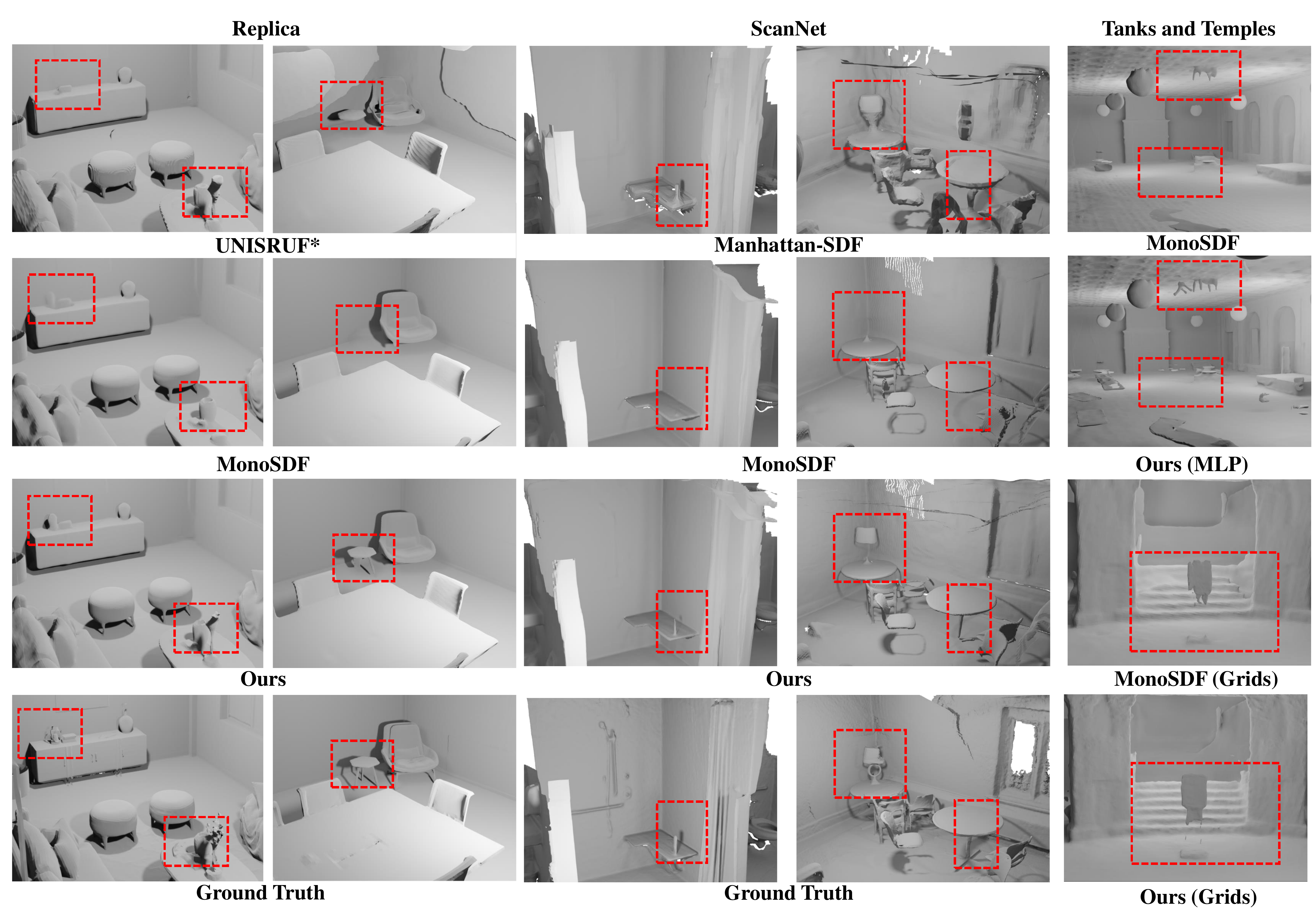}
    \caption{\textbf{Reconstruction results on representative datasets} of Replica (left), ScanNet (middle), and Tanks and Temples (right). The ground truth is presented on the bottom-most. Red boxes in sub-figures highlight those areas where distinctive differences can be observed.}
    \label{fig: All Results}
    \vspace{-12pt}
\end{figure*}

{\noindent\textbf{Results on ScanNet Dataset.}
We conducted a comparative analysis of our proposed approach against existing implicit reconstruction methods, including Manhattan-SDF~\cite{guo2022neural}, NeuRIS~\cite{wang2022neuris} and MonoSDF~\cite{yu2022monosdf} using the ScanNet dataset. As revealed in Table~\ref{tab: scannet}, our proposed method outperforms state-of-the-art methods, with a significant increase in F-score by \textbf{4.6}. Additionally, in terms of Recall, our method substantially outperforms MonoSDF by \textbf{7.9} without needing extra data. 
{Overall, our approach performs on par with the state-of-art methods in  ``Acc'', ``Chamfer-$\mathcal{L}_1$(C-$\mathcal{L}_1$)'' and ``Prec'' and obtain notable performance gains in ``Comp'', ``Recall'' and ``F-score''. 
This is because these metrics ("Comp" and "Recall") are better metrics in evaluating how complete and accurate in capturing the shape and details of the scene being reconstructed.
} 
 Further, Fig.~\ref{fig: All Results} reveals that our method can attain more complete reconstructions with details and for low pixel intensities regions. 
}
\begin{table}[t]
  \centering
  \tablestyle{3.4pt}{1}
  \resizebox{0.99\linewidth}{!}{\begin{tabular}{l|ccccc}
    \hline
                       & Auditorium & Ballroom& Courtroom& Museum& Mean\\
    \hline
    MonoSDF    & 3.09 & 2.47 & 10.00  & 5.10    & 5.165\\
    \textbf{Ours}      & \textbf{5.22} & \textbf{5.42} & \textbf{13.99}  & \textbf{8.59}    & \textbf{8.305}\\
    \hline
    MonoSDF*    & 3.17 & 3.70 & 13.75  & 5.68    & 6.58\\
    \textbf{Ours*}      & \textbf{6.19} & \textbf{7.33} & \textbf{19.80}  & \textbf{11.85}    & \textbf{11.295}\\
    \hline
  \end{tabular}}
  \vspace{2pt}
  \caption{Quantitative assessments of the proposed model against Monosdf on the Tanks and Temples dataset. The evaluation metrics for the Tanks and Temples dataset are F-score. * means that the hash-grid structure is adopted.}
  \label{tab: all in tnt}
  \vspace{-17pt}
\end{table}

{\noindent\textbf{Results on Tanks and Temples Dataset.}
For challenging large-scale indoor scenes, we conduct experiments on the advanced Tanks and Temples dataset~\cite{Knapitsch2017}, which features more complicated structures.
As alternative methods of neural reconstruction from images are not assessed on this dataset, we implement the best-performing method MonoSDF and compare with it. 
Thus, the MonoSDF and two versions of our method are implemented. Specifically, one adopts the pure MLP architecture while the other uses the hash grids as the input representation. 
The quantitative assessments (Table~\ref{tab: all in tnt}) reveal that our method shows better performance on this dataset, regardless of whether an MLP or hash-grid~\cite{mueller2022instant} structure is used. And our hybrid representation exhibits excellent generalization abilities across different implicit structures. Overall, our compelling experimental results on the Tanks and Temples dataset further validate the robustness and versatility of the proposed method in reconstructing complex and challenging indoor scenes.}

\begin{table}[h]
  \centering
  \vspace{-5pt}
  \resizebox{0.99\linewidth}{!}{
      \footnotesize
    \begin{tabular}{l|ccc}
    \hline
    Method & Normal C.$\uparrow$  &Chamfer-$\mathcal{L}_1$ $\downarrow$  &F-score $\uparrow$ \\
    \hline
    UNISURF$^\dagger$          &90.96 &4.93 &78.99 \\
    MonoSDF$^\dagger$          &92.11 &2.94 &86.18 \\
    \textbf{Ours}$^\dagger$  &\textbf{93.35} &\textbf{2.58} &\textbf{91.25}\\
    \hline
  \end{tabular}}
  \vspace{2pt}
  \caption{Quantitative assessments of the proposed model against prior works on the Replica dataset. Herein, $^\dagger$ indicates the use of geometry priors as supervision signals.}
  \label{tab:all in replica}
  \vspace{-5pt}
\end{table}

\noindent\textbf{Results on Replica Dataset.}
Quantitative assessment results on the Replica dataset are presented in Table~\ref{tab:all in replica}. 
Ours significantly surpasses existing state-of-the-art neural rendering methods. 
The results reveal that the SDF-based representation outperforms the occupancy-based ones (\ie UNISURF*).
This is because the SDF usually enforces constraints on the distribution of the entire scene, benefiting to suppressing the occurrence of floaters or unnecessary structures in occupancy-based representation.
Notably, our {Occ-SDF Hybrid} method can constrain the distribution of the entire scene with SDF representation meanwhile exploiting the occupancy representation to resolve thin structures and small objects. 
{Qualitative comparisons are shown in Fig.~\ref{fig: All Results}}.

\subsection{Ablation Study}
\begin{table}[h]
  \centering
  \resizebox{0.9\linewidth}{!}{
     \footnotesize
     \begin{tabular}{l|ccc}
     \hline
     & Normal C.$\uparrow$ & Chamfer-$\mathcal{L}_1$ $\downarrow$ & F-score $\uparrow$
      \\
     \hline
    MonoSDF                    & 92.11 & 2.94 & 86.18 \\
    \; + feature     & 93.01 & 2.64 & 91.01 \\
    \; + hybrid & 93.22 & 2.77 & 90.24 \\
    \; full model  &\textbf{93.35} &\textbf{2.58} &\textbf{91.25}\\
    \hline
  \end{tabular}}
  \vspace{2pt}
  \caption{Ablation study on the Replica dataset~\cite{replica19arxiv}, where we progressively add different constraints to assess their impacts. The MonoSDF~\cite{yu2022monosdf} is set as the baseline model.}
  \label{tab: ablation}
  \vspace{-5pt}
\end{table}
\noindent We conduct ablation studies on the Replica dataset as it provides ground-truth geometry. Four different configurations are investigated to train our model, including (1) MonoSDF with MLP settings (MonoSDF-MLP); (2) MonoSDF-MLP with our feature-based rendering formula; (3) MonoSDF-MLP with our hybrid representation; (4) MonoSDF-MLP with both the feature-based rendering formula and hybrid representation scheme (Full model). 

Table~\ref{tab: ablation} shows that all metrics are improved when using the feature rendering to reconstruct this scene. 
Our proposed feature rendering scheme addresses the difficulties in reconstructing areas of low intensities, resulting in better results. 
On the other hand, the hybrid representation also leads to significant improvements in all metrics. Notably, it improves the completeness of small objects and thin structures, as evidenced by the results in Fig.~\ref{fig: All Results}. 
By leveraging both components, our model achieves an overall improvement of \textbf{5.07} in F-score, along with improved normal consistency and Chamfer-$\mathcal{L}_1$. 
We attribute this success to our feature-based color rendering formula and our hybrid representation, which addresses the color-bias issue in optimization and difficulties in reconstructing detailed structures.
The visualization results in Fig.~\ref{fig: All Results} show our model's excellent reconstruction performance, especially in low-intensity areas and detailed structures. 
{We will add more ablation studies and visualize results in the supplement.}

\vspace{-6pt}
\section{Conclusion}
We have analyzed the constraints present in current neural scene representation techniques with geometry priors, and have identified issues in their ability to reconstruct detailed structures due to a biased optimization towards high color intensities and the complex SDF distribution. As a result, we have developed a feature rendering scheme that balances color regions and have implemented a hybrid representation to address the limitations of the SDF distribution. Our approach has demonstrated the successful reconstruction of room scenes with a high-fidelity surface, including small objects, detailed structures, and low-intensity pixel regions. We envision our results inspire further research on improving neural scene representation for accurate and large-scale surface reconstruction.

{\small
\bibliographystyle{ieee_fullname}
\bibliography{egbib}
}

\end{document}